\documentclass{ieeeaccess}

\usepackage{cite}
\usepackage{amsmath,amssymb,amsfonts}
\usepackage{algorithmic}
\usepackage{graphicx,subcaption}
\usepackage{textcomp}
\usepackage{booktabs,multirow}
\usepackage{hyperref}

\newcommand{\highlighting}{black}
    
\begin{document}

	\history{Received 7 November 2023, accepted 19 January 2024, date of publication 24 January 2024, date of current version 1 February 2024.}
	
	\doi{10.1109/ACCESS.2024.3357887}

	\title{Explicit Feature Interaction-Aware\\Graph Neural Network}
	
	\author{
		\uppercase{
			Minkyu Kim\authorrefmark{1},
			Hyun-Soo Choi\authorrefmark{2},
			and Jinho Kim\authorrefmark{3}, \IEEEmembership{Member, IEEE}
		}
	}

	\address[1]{%
		Department of Research and Development,
		Ziovision Company Ltd.,
		Chuncheon 24341,
		Republic of Korea
	}
	\address[2]{%
		Department of Computer Science and Engineering,
		Seoul National University of Science and Technology,
		Seoul 01811,
		Republic of Korea
	}
	\address[3]{%
		Department of Computer Science and Engineering,
		Kangwon National University,
		Chuncheon 24341,
		Republic of Korea
	}

	\tfootnote{This work was supported by the National Research Foundation of Korea(NRF) grant funded by the Korea government(MSIT) (No. 2021R1F1A1059255).}

	\markboth
	{Minkyu Kim \headeretal: Explicit Feature Interaction-aware Graph Neural Network}
	{Minkyu Kim \headeretal: Explicit Feature Interaction-aware Graph Neural Network}

	\corresp{Corresponding author: Hyun-Soo Choi (e-mail: choi.hyunsoo@seoultech.ac.kr), and Jinho Kim (e-mail: jhkim@kangwon.ac.kr).}

	\begin{abstract}
		Graph neural networks (GNNs) are powerful tools for handling graph-structured data. However, their design often limits them to learning only higher-order feature interactions, leaving low-order feature interactions overlooked. To address this problem, we introduce a novel GNN method called explicit feature interaction-aware graph neural network (EFI-GNN). Unlike conventional GNNs, EFI-GNN is a multilayer linear network designed to model arbitrary-order feature interactions explicitly within graphs. To validate the efficacy of EFI-GNN, we conduct experiments using various datasets. The experimental results demonstrate that EFI-GNN has competitive performance with existing GNNs, and when a GNN is jointly trained with EFI-GNN, predictive performance sees an improvement. Furthermore, the predictions made by EFI-GNN are interpretable, owing to its linear construction. The source code of EFI-GNN is available at \url{https://github.com/gim4855744/EFI-GNN}.
	\end{abstract}


	\titlepgskip=-21pt

	\maketitle
	
    \section{Introduction}

    Graphs are ubiquitous in the real world. Molecules, social networks, citation networks, and natural languages are representative examples. Directly analyzing such complex graphs is challenging, and embedding graphs(or nodes) into real vectors is necessary for applying machine learning techniques. As a result, graph-representation learning has emerged as a crucial task in recent years.

    Graph neural networks (GNNs) are attracting great attention across various domains due to their remarkable ability to handle graph-structured data \cite{fan2019graph,huang2018adaptive,kipf2017semi}. GNNs aggregate input features and the representations of neighboring nodes in nonlinear ways utilizing an activation function. This approach enables the implicit learning of higher-order feature interactions, which can be advantageous in capturing intricate patterns. However, some valuable patterns can be captured from low-order feature interactions (e.g., 1st-, 2nd-, or 3rd-order interactions) \cite{cheng2016wide,guo2017deepfm,lian2018xdeepfm}. By focusing solely on higher-order feature interactions, GNNs might miss essential patterns, potentially resulting in a degradation of predictive performance.
    
    Explicitly learning feature interactions- a way humans can understand how features are combined and identify the influence of these interactions on output values- has been widely studied \cite{cheng2016wide,guo2017deepfm,wang2017deep,lian2018xdeepfm}. To learn low-order feature interactions, existing methods employ linear regression \cite{cheng2016wide}, factorization \cite{rendle2010factorization,guo2017deepfm}, or feature crossing \cite{wang2017deep,lian2018xdeepfm,kim2020combining}. They have also shown that combining low-order and higher-order feature interactions can enhance predictive performance in tasks like recommendations \cite{guo2017deepfm,lian2018xdeepfm} and regression \cite{kim2020combining}. However, to the best of our knowledge, there are no explicit feature interaction methods that can directly handle graphs. In addition, although explicit methods are inherently interpretable, their interpretability has been neglected in the previous literature.

    To overcome the aforementioned problem, we propose a novel GNN method named explicit feature interaction-aware graph neural network (EFI-GNN), which can explicitly learn arbitrary-order feature interactions on a graph. EFI-GNN is a multilayer linear network. Specifically, EFI-GNN performs graph convolution \cite{kipf2017semi} without an activation function and multiplies the sum of 1st-order features across each layer, which we refer to as feature crossing. Therefore, the interaction order of EFI-GNN escalates with the increase in its number of layers. For instance, 2- and 4-layer EFI-GNNs can learn 2nd- and 4th-order feature interactions, respectively. In addition, EFI-GNN is inherently interpretable due to its linearity.

    We conduct experiments using various citation network datasets to validate the effectiveness of EFI-GNN. The experimental results demonstrate that EFI-GNN has competitive predictive performance with existing GNNs. Moreover, jointly training an existing GNN with EFI-GNN can improve predictive performance. To determine the efficacy of learning multiple feature interactions in graphs, we conduct further experiments across various joint-learning scenarios using open graph benchmark (OGB) datasets. The experimental results demonstrate that combining multiple feature interactions leads to performance improvements. This implies that sophisticated techniques, previously proposed for learning feature interactions to improve recommendation or regression performances, may be potent for graph applications. In addition, to interpret the prediction of EFI-GNN, we visualize 1st- and 2nd-order feature interactions as heatmaps.

    \section{Related Works}

    \subsection{Graph Neural Networks}

        GNNs are a special form of deep neural networks tailored for graph-structured data. Vanilla graph neural network (VanillaGNN) \cite{scarselli2008graph} extends recurrent neural networks (RNNs) \cite{rumelhart1986learning} to make them applicable to more general graphs, like undirected cyclic graphs. Gated GNN (GGNN) \cite{li2016gated} improves VanillaGNN by using gated recurrent unit (GRU) \cite{cho2014learning}. These RNN-based GNNs can be classified into Recurrent GNNs (RGNNs). Planetoid \cite{yang2016revisiting} is a semi-supervised node embedding framework trained to predict node class labels and subgraph contexts simultaneously through a neural network. Recently, graph convolutional network (GCN) \cite{kipf2017semi} and its variants \cite{hamilton2017inductive,velivckovic2018graph} have achieved great success in the node embedding task. One problem with spectral methods like GCN is over-smoothing. To alleviate the over-smoothing problem and train a deep GCN, several methods utilizing residual connections have been proposed \cite{li2019deepgcns,li2020deepergcn}. Jumping knowledge network (JKNet) \cite{xu2018representation} concatenates all hidden layers of GCN to make predictions. This strategy can alleviate the over-smoothing problem. Subgraph sampling techniques for fast training and inference of GCN have been proposed \cite{chen2018fastgcn,huang2018adaptive}. Simple graph convolutional network (SGC) \cite{wu2019simplifying} is a straightforward and linear version of GCN using powers of the adjacency matrix. Link prediction methods for bipartite graphs utilizing GNNs have been proposed \cite{berg2017graph,zhang2020inductive}. Graph transformer \cite{yun2019graph} automatically generates meta-paths and effectively deals with heterogeneous graphs. All the aforementioned methods except SGC implicitly aggregate feature information. Thus, they only learn higher-order feature interactions and cannot capture valuable patterns occurring in low-order feature interactions.
	
    \subsection{Feature Interaction Methods}

        Feature interactions refer to the combined effects of different features on a given output value. Such interactions play a crucial role in improving the performances of machine learning models \cite{rendle2010factorization,guo2017deepfm}. Traditionally, handcrafted features have been widely used to learn feature interactions. However, generating handcrafted features requires domain expertise and is time-consuming. Recently, various methods have been proposed for learning feature interactions without handcrafting. One example is factorization machine (FM) \cite{rendle2010factorization}, which combines linear regression and factorization methods to learn 1st- and 2nd-order feature interactions simultaneously. Similarly, Wide \& deep learning \cite{cheng2016wide} combines linear regression and deep neural network (DNN) to learn 1st- and higher-order feature interactions simultaneously. DeepFM \cite{guo2017deepfm} is a joint learning method of FM and DNN. The aforementioned models learn only bounded-order feature interactions. To overcome this problem, CrossNet \cite{wang2017deep} has been proposed. It can learn arbitrary-order feature interactions based on its number of layers. Compressed interaction network (CIN) \cite{lian2018xdeepfm} and the explicit component of extreme interaction network (XIN) \cite{kim2020combining} have extended CrossNet to vector-wise operations.
    
    \subsection{Feature Interactions via Graphs}

        In recent years, several methods have been proposed for learning feature interactions through GNNs \cite{li2019fi,li2021graphfm,liu2021gcn,zheng2021graph}. These methods designate features as nodes and employ GNNs to learn edge weights within the feature graph. The learned edge weights indicate the interactions between different features. However, they cannot handle graph-structured data. FI-GNN \cite{ding2019feature} is a joint learning method of GNN and the feature factorization module. However, the feature factorization module cannot apply directly to graph-structured data and only learns 2nd-order feature interactions. In contrast, our proposed EFI-GNN can directly handle graph-structured data and learn arbitrary-order feature interactions.

    \section{EFI-GNN}

    \begin{figure*}[t]
    \centering
    \includegraphics[width=0.65\textwidth]{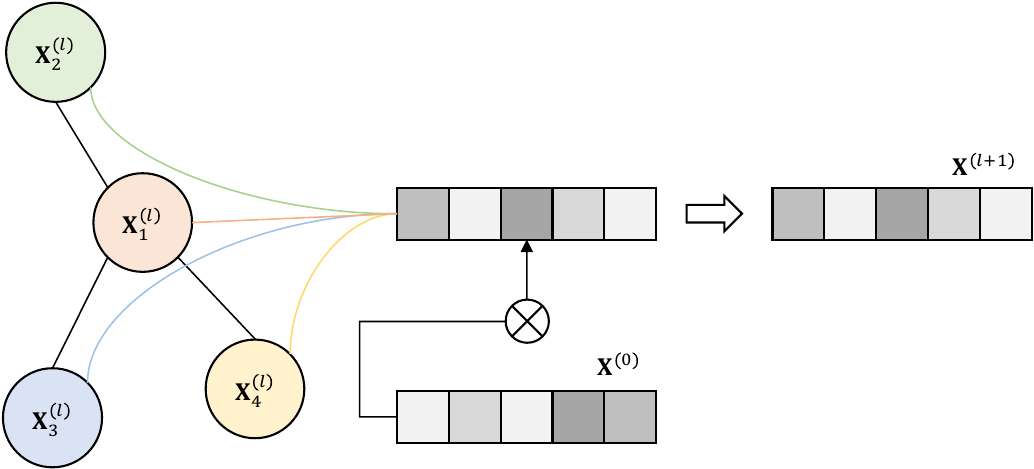}
    \caption{EFI-GNN layer}
    \label{fig:layer}
\end{figure*}

    \subsection{Preliminaries}

        \subsubsection{Feature Interaction}

            Feature interactions are the influences of features to model outcomes. 1st-order feature interactions indicate the influences of single features. Similarly, 2nd- and 3rd-order feature interactions indicate the influences of 2nd- and 3rd-order features (e.g., user $\times$ movie, user $\times$ movie $\times$ date). More formally, feature interaction can be defined as follows:

            \noindent
            \noindent \textbf{Definition 1.}
            If a function $h(\cdot)$ holds the following condition, there are interactions between features $\textbf{x}$ \cite{tsang2020does}.
            \begin{equation}
                h \left( x_1 + x_2 + ... + x_m \right) \neq \sum_{i=1}^{m} h \left( x_i \right).
            \end{equation}
            All deep learning models always hold the above condition due to their non-linearity. Therefore, deep learning models always and only learn interactions between all features (i.e., higher-order feature interactions).
	
    \subsection{EFI-GNN Layer}

        EFI-GNN has a multilayer architecture but is quite different from existing GNNs. Each layer in EFI-GNN has a feature-crossing term, which multiplies the sum of 1st-order features instead of an activation function. In EFI-GNN, the sum of 1st-order features is defined as follows:
        \begin{equation}
            \textbf{X}^{(1)} = \textbf{X}^{(0)} \textbf{W}^{(1)}, \label{eq:x0}
        \end{equation}
        where $\textbf{X}^{(0)} \in \mathbb{R}^{N \times M}$ is the raw feature matrix, $\textbf{W}^{(1)} \in \mathbb{R}^{M \times K}$ is the trainable weight matrix, $N$ is the number of nodes in the graph, $M$ is the number of features, and $K$ is the projection size. An EFI-GNN layer is defined as follows:
        \begin{equation}
            \textbf{X}^{(l)} = \hat{\textbf{A}} \textbf{X}^{(l-1)} \textbf{W}^{(l)} \odot \textbf{X}^{(1)}, \label{eq:xl}
        \end{equation}
        where $\hat{\textbf{A}} = \tilde{\textbf{D}}^{-1/2} \tilde{\textbf{A}} \tilde{\textbf{D}}^{-1/2}$, $\tilde{\textbf{A}} = \textbf{A} + \textbf{I}$, $\textbf{A} \in \mathbb{R}^{N \times N}$ is the adjacency matrix, $\tilde{\textbf{D}} \in \mathbb{R}^{N \times N}$ is the degree matrix of $\tilde{\textbf{A}}$, $\textbf{W}^{(l)} \in \mathbb{R}^{K \times K}$ is the trainable weight matrix in $l^{th}$ layer, and $\odot$ indicates the Hadamard product. The adjacency matrix $\hat{\textbf{A}}$ is multiplied by the feature matrix $\textbf{X}^{(l-1)}$ to aggregate the neighbor nodes' information, and the sum of 1st-order features is multiplied in each layer to learn high-order feature interactions. Therefore, the order of feature interactions that EFI-GNN learns gradually escalates with the number of layers increases. Fig.~\ref{fig:layer} depicts the architecture of an EFI-GNN layer.

        Since each layer of EFI-GNN captures different feature interactions, the output layer of EFI-GNN leverages these rich feature interactions to enhance its predictive performance. The final output layer of EFI-GNN is defined as follows:
        \begin{equation}
            \hat{\textbf{Y}} = \left[ \textbf{X}^{(1)} \parallel \textbf{X}^{(2)} \parallel ... \parallel \textbf{X}^{(L)} \right] \textbf{W}^{(\text{out})},
        \end{equation}
        where $\parallel$ is the concatenation operator, $L$ is the number of layers, $\textbf{W}^{(\text{out})} \in \mathbb{R}^{K \cdot L \times O}$ is the trainable output matrix, and $O$ is the output size.

    \begin{figure}
    \centering
    \includegraphics[width=0.9\linewidth]{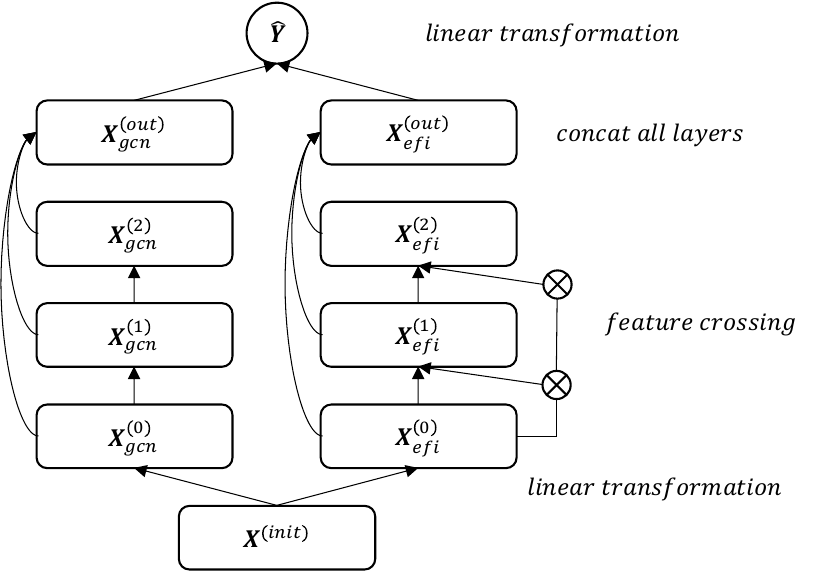}
    \caption{The architecture of GCN \& EFI-GNN.}
    \label{fig:gcn_cross_arc}
\end{figure}
	\begin{table*}[t]

    \centering
    
    \begin{tabular}{lccccccc}
    
        \toprule
        
        & \textbf{Cora}
        & \textbf{CiteSeer}
        & \textbf{PubMed}
        & \textcolor{\highlighting}{\textbf{Amazon}}
        & \textcolor{\highlighting}{\textbf{Actor}}
        & \textbf{ogbn-arxiv}
        & \textbf{ogbn-mag} \\
        
        \midrule
        
        \# of Nodes
            & 2,708
            & 3,327
            & 19,717
            & \textcolor{\highlighting}{13,752}
            & \textcolor{\highlighting}{7,600}
            & 169,343
            & 1,939,743 \\
        \# of Edges
            & 5,429
            & 4,732
            & 44,338
            & \textcolor{\highlighting}{491,722}
            & \textcolor{\highlighting}{30,019}
            & 1,166,243
            & 21,111,007 \\
        \# of Features
            & 1,433
            & 3,703
            & 500
            & \textcolor{\highlighting}{767}
            & \textcolor{\highlighting}{932}
            & 128
            & 128 \\
        \# of Classes
            & 7
            & 6
            & 3
            & \textcolor{\highlighting}{10}
            & \textcolor{\highlighting}{5}
            & 40
            & 349 \\

        \midrule
        
        \# of Training Nodes
            & 1,208
            & 1,827
            & 18,217
            & \textcolor{\highlighting}{8,252}
            & \textcolor{\highlighting}{3,648}
            & 90,941
            & 629,571 \\
        \# of Validation Nodes
            & 500
            & 500
            & 500
            & \textcolor{\highlighting}{2,750}
            & \textcolor{\highlighting}{2,432}
            & 29,799
            & 64,879 \\
        \# of Test Nodes
            & 1,000
            & 1,000
            & 1,000
            & \textcolor{\highlighting}{2,750}
            & \textcolor{\highlighting}{1,520}
            & 48,603
            & 41,939 \\

        \midrule
        
        Graph Type
            & undirected
            & undirected
            & undirected
            & \textcolor{\highlighting}{undirected}
            & \textcolor{\highlighting}{undirected}
            & directed
            & directed \\
        Feature Type
            & zero/one
            & zero/one
            & tf/idf
            & \textcolor{\highlighting}{zero/one}
            & \textcolor{\highlighting}{zero/one}
            & word2vec
            & word2vec \\
        
        \bottomrule
        
    \end{tabular}
    
    \caption{Dataset statistics}
    
    \label{tbl:dataset_statistics}
    
\end{table*}

	\begin{table*}[t!]

    \centering
    
    \begin{tabular}{ccccccc}
    
        \toprule
        
        & \textbf{Cora}
        & \textbf{CiteSeer}
        & \textbf{PubMed}
        & \textcolor{\highlighting}{\textbf{Amazon}}
        & \textcolor{\highlighting}{\textbf{Actor}}
        & \textbf{OGB} \\
        
        \midrule
        
        \# of GNN Layers
            & 3 & 3 & 3 & \textcolor{\highlighting}{3} & \textcolor{\highlighting}{3} & 1 \\
        \# of EFI-GNN Layers
            & 2 & 2 & 2 & \textcolor{\highlighting}{2} & \textcolor{\highlighting}{2} & 1 \\
        \# of Units
            & 128 & 128 & 1024 & \textcolor{\highlighting}{128} & \textcolor{\highlighting}{128} & 128 \\
        Learning Rate
            & 1e-3 & 1e-3 & 1e-3 & \textcolor{\highlighting}{1e-3} & \textcolor{\highlighting}{1e-3} & 1e-2 \\
        L2 Penalty
            & 1e-2 & 1e-2 & 1e-3 & \textcolor{\highlighting}{1e-2} & \textcolor{\highlighting}{1e-2} & No \\
        Dropout
            & 0.9 & 0.9 & 0.85 & \textcolor{\highlighting}{0.9} & \textcolor{\highlighting}{0.9} & 0.3 \\
        Activation
            & LeakyReLU & LeakyReLU & LeakyReLU & \textcolor{\highlighting}{LeakyReLU} & \textcolor{\highlighting}{LeakyReLU} & LeakyReLU \\
        Residual Connection
            & No & No & Dense & \textcolor{\highlighting}{No} & \textcolor{\highlighting}{No} & No \\
        Batch Normalization
            & No & No & Yes & \textcolor{\highlighting}{No} & \textcolor{\highlighting}{No} & Yes \\
        
        \bottomrule
        
    \end{tabular}
    
    \caption{Hyperparameter settings for the experimental datasets}
    
    \label{tbl:hparam1}
    
\end{table*}

    \subsection{EFI-GNN Layer Analysis}
        
        To reveal how the EFI-GNN layer captures feature interactions, we present the vector-level representation of the layer. In (\ref{eq:x0}), $\textbf{X}^{(1)}$ represents the linear transformation of the input features. In other words, $\textbf{X}^{(1)}$ indicates the weighted sum of 1st-order features. Therefore, we can express (\ref{eq:x0}) at the vector level as:
        \begin{align}
            \begin{split}
                \textbf{x}_{v}^{(1)}
                &= \left( x_{v,1}^{(0)} \textbf{w}^{(1)}_{1} + x_{v,2}^{(0)} \textbf{w}^{(1)}_{2} + ... + x_{v,M}^{(0)} \textbf{w}^{(1)}_{M} \right) \\
                &= \sum_{m=1}^{M} x_{v,m}^{(0)} \textbf{w}^{(1)}_{m},
            \end{split}
            \label{eq:x0v}
        \end{align}
        where $v$ denotes the node index. (\ref{eq:x0v}) shows that $\textbf{x}_{v}^{(1)}$ is the weighted sum of the input features of node $v$. In (\ref{eq:xl}), $\textbf{X}^{(l)}$ stands for the linear transformation of $\textbf{X}^{(l-1)}$, achieved through neighbor aggregation and feature-crossing. Therefore, (\ref{eq:xl}) can be expressed at the vector level as:
        \begin{align}
            \begin{split}
                \textbf{x}_{v}^{(l)}
                &= \left( \hat{\textbf{A}} \textbf{X}^{(l-1)} \textbf{W}^{(l)} \right)_{v} \odot \textbf{x}_{v}^{(1)} \\
                &= \sum_{i \in \mathcal{N}(v)} \left( x_{i,1}^{(l-1)} \textbf{w}^{(l)}_{1} + x_{i,2}^{(l-1)} \textbf{w}^{(l)}_{2} + ... + x_{i,K}^{(l-1)} \textbf{w}^{(l)}_{K} \right) \\
                & \hspace{29pt} \odot \left( x_{v,1}^{(0)} \textbf{w}^{(1)}_{1} + x_{v,2}^{(0)} \textbf{w}^{(1)}_{2} + ... + x_{v,M}^{(0)} \textbf{w}^{(1)}_{M} \right) \\
                &= \sum_{i \in \mathcal{N}(v)} \sum_{k=1}^{K} x_{i,k}^{(l-1)} \textbf{w}^{(l)}_{k} \odot \sum_{m=1}^{M} x_{v,m}^{(0)} \textbf{w}^{(1)}_{m}.
            \end{split}
            \label{eq:xlv}
        \end{align}
        In (\ref{eq:xlv}), the previous layer, $\sum_{i \in \mathcal{N}(v)} \sum_{k=1}^{K} x_{i,k}^{(l-1)} \textbf{w}^{(l)}_{k}$, is multiplied by the sum of 1st-order features, $\sum_{m=1}^{M} x_{v,m}^{(0)} \textbf{w}^{(1)}_{m}$. This allows EFI-GNN layers to learn high-order feature interactions gradually.

        As highlighted above, feature crossing is a pivotal component of our EFI-GNN. Standard linear models possess limited expressive power regardless of their number of layers. This limitation arises because they can only learn 1st-order feature interactions and linear patterns. However, linear models with feature crossing can learn high-order feature interactions, and this strategy allows linear models to partially overcome the limitation \cite{wang2017deep,lian2018xdeepfm}.

    \begin{table*}[t!]

    \centering
    
    \begin{tabular}{clccccc}
        
        \toprule
        
        & & \textbf{Cora}
        & \textbf{CiteSeer}
        & \textbf{PubMed}
        & \textcolor{\highlighting}{\textbf{Amazon}}
        & \textcolor{\highlighting}{\textbf{Actor}} \\
        
        \midrule
        
        \multirow{5}{*}{reported}
        & GCN \cite{kipf2017semi}
            & 86.64\%
            & 79.34\%
            & 90.22\%
            & \textcolor{\highlighting}{N/A}
            & \textcolor{\highlighting}{N/A} \\
        & FastGCN \cite{chen2018fastgcn}
            & 85.00\%
            & 77.60\%
            & 88.00\%
            & \textcolor{\highlighting}{N/A}
            & \textcolor{\highlighting}{N/A} \\
        & GraphSage \cite{hamilton2017inductive}
            & 82.20\%
            & 71.40\%
            & 87.10\%
            & \textcolor{\highlighting}{N/A}
            & \textcolor{\highlighting}{N/A} \\
        & ASGCN \cite{huang2018adaptive}
            & 87.44$\pm$0.34\%
            & 79.66$\pm$0.18\%
            & 90.60$\pm$0.16\%
            & \textcolor{\highlighting}{N/A}
            & \textcolor{\highlighting}{N/A} \\
        & GCNII \cite{chen2020simple}
            & 88.49\%
            & 77.08\%
            & 89.57\%
            & \textcolor{\highlighting}{N/A}
            & \textcolor{\highlighting}{N/A} \\
        
        \midrule
        
        \multirow{12}{*}{our}
        
        & EFI-GNN
            & 87.95$\pm$0.53\%
            & 78.45$\pm$0.57\%
            & 89.85$\pm$0.22\%
            & \textcolor{\highlighting}{88.43$\pm$0.12\%}
            & \textcolor{\highlighting}{30.14$\pm$0.19\%} \\

        \cmidrule{2-7}
        
        & GCN\cite{kipf2017semi}
            & 87.88$\pm$0.20\%
            & 79.23$\pm$0.31\%
            & 90.07$\pm$0.40\%
            & \textcolor{\highlighting}{88.75$\pm$0.10\%}
            & \textcolor{\highlighting}{29.72$\pm$0.26\%} \\
        & \hfill$+$ EFI-GNN
            & \textbf{88.60$\pm$0.21\%}
            & \textbf{79.80$\pm$0.20\%}
            & \textbf{90.87$\pm$0.23\%}
            & \textcolor{\highlighting}{\textbf{88.98$\pm$0.13\%}}
            & \textcolor{\highlighting}{\textbf{30.28$\pm$0.09\%}} \\

        \cmidrule{2-7}

        & GAT\cite{velivckovic2018graph}
            & 86.78$\pm$0.22\%
            & 78.92$\pm$0.31\%
            & 88.64$\pm$0.32\%
            & \textcolor{\highlighting}{83.27$\pm$1.81\%}
            & \textcolor{\highlighting}{29.74$\pm$0.21\%} \\
        & \hfill $+$ EFI-GNN
            & 87.50$\pm$0.24\%
            & 80.54$\pm$0.18\%
            & 89.82$\pm$0.15\%
            & \textcolor{\highlighting}{88.12$\pm$0.79\%}
            & \textcolor{\highlighting}{30.28$\pm$0.16\%} \\

        \cmidrule{2-7}
        
        & GCNII\cite{chen2020simple}
            & 88.41$\pm$0.32\%
            & 79.42$\pm$0.23\%
            & 90.34$\pm$0.21\%
            & \textcolor{\highlighting}{88.59$\pm$0.10\%}
            & \textcolor{\highlighting}{29.97$\pm$0.27\%} \\
        & \hfill$+$ EFI-GNN
            & \textbf{88.77$\pm$0.28\%}
            & \textbf{79.54$\pm$0.37\%}
            & \textbf{90.41$\pm$0.34\%}
            & \textcolor{\highlighting}{88.76$\pm$0.17\%}
            & \textcolor{\highlighting}{30.00$\pm$0.23\%} \\

        \cmidrule{2-7}

        & GATv2\cite{brody2022attentive}
            & 86.14$\pm$0.18\%
            & 78.84$\pm$0.30\%
            & 89.12$\pm$0.22\%
            & \textcolor{\highlighting}{85.50$\pm$2.45\%}
            & \textcolor{\highlighting}{30.20$\pm$0.19\%} \\
        & \hfill $+$ EFI-GNN
            & 86.88$\pm$0.60\%
            & 80.50$\pm$0.19\%
            & 90.02$\pm$0.32\%
            & \textcolor{\highlighting}{87.61$\pm$0.63\%}
            & \textcolor{\highlighting}{30.12$\pm$0.36\%} \\

        \cmidrule{2-7}

        & ASDGN\cite{gravina2023anti}
            & 85.68$\pm$0.29\%
            & 78.42$\pm$0.19\%
            & 88.68$\pm$0.36\%
            & \textcolor{\highlighting}{82.81$\pm$0.96\%}
            & \textcolor{\highlighting}{29.68$\pm$0.28\%} \\
        & \hfill $+$ EFI-GNN
            & 87.86$\pm$0.45\%
            & 79.60$\pm$0.07\%
            & 89.58$\pm$0.24\%
            & \textcolor{\highlighting}{88.47$\pm$0.12\%}
            & \textcolor{\highlighting}{30.20$\pm$0.32\%} \\
            
        \bottomrule
        
    \end{tabular}
    
    \caption{Classification accuracies on the Cora, CiteSeer, PubMed, Amazon, and Actor datasets}
    
    \label{tbl:cite_acc}
    
\end{table*}

    \begin{table}[t!]

    \centering
    
    \begin{tabular}{ccc}
    
        \toprule
        
        & \textbf{ogbn-arxiv} & \textbf{ogbn-mag} \\
        
        \midrule
        
        GCN$^*$
            & 56.61$\pm$0.15\%
            & 26.35$\pm$0.25\% \\
            
        EFI-GNN$^*$
            & 56.82$\pm$0.26\%
            & 25.82$\pm$0.29\% \\
            
        \midrule
        
        GCN
            & 56.94$\pm$0.07\%
            & 26.58$\pm$0.17\% \\

        GAT
            & 57.20$\pm$0.20\%
            & 25.75$\pm$0.20\% \\

        GATv2
            & 56.26$\pm$0.19\%
            & 25.76$\pm$0.19\% \\

        ASDGN
            & 56.72$\pm$0.30\%
            & 25.78$\pm$0.11\% \\
            
        EFI-GNN
            & 57.37$\pm$0.28\%
            & 26.25$\pm$0.24\% \\
        
        \midrule
        
        GCN \& GCN
            & 57.40$\pm$0.15\%
            & 26.58$\pm$0.38\% \\
            
        EFI-GNN \& EFI-GNN
            & 57.45$\pm$0.19\%
            & 26.25$\pm$0.19\% \\
        
        GCN \& EFI-GNN
            & \textbf{57.96$\pm$0.13\%}
            & \textbf{26.79$\pm$0.22\%} \\

        GAT \& EFI-GNN
            & 57.25$\pm$0.26\%
            & 25.87$\pm$0.18\% \\

        GATv2 \& EFI-GNN
            & 56.56$\pm$0.11\%
            & 26.00$\pm$0.27\% \\

        ASDGN \& EFI-GNN
            & 57.45$\pm$0.26\%
            & 26.38$\pm$0.10\% \\
        
        \bottomrule
        
    \end{tabular}
    
    \caption{Classification accuracies on the OGB datasets}
    
    \label{tbl:ogb_acc}
    
\end{table}

    \subsection{Interpretability}

        If an EFI-GNN layer is directly connected to the output layer, we can obtain the influences of features within that layer, leveraging the intrinsically linear nature of EFI-GNN. We describe the procedure for obtaining feature influences below. The influence of a 1st-order feature is defined as follows:
        \begin{gather}
            \textbf{a}_{n,i}^{(1)} = x_{n,i}^{(0)} \textbf{w}_{i}^{(1)}, \\
            e_{c,n,i}^{(1)} = \textbf{a}_{n,i}^{(1)} \textbf{w}_{:k,c}^{(\text{out})},
        \end{gather}
        where $\textbf{a}^{(1)}_{n,i} \in \mathbb{R}^{k}$ is the representation vector for the 1st-order feature $i$ of the node $n$, $\textbf{w}^{(1)}_{i} \in \mathbb{R}^{k}$ denotes the weight vector for feature $i$, $\textbf{w}^{(\text{out})}_{:u,c} \in \mathbb{R}^{k}$ is the output weight vector corresponding to the class $c$, $k$ denotes the number of units in the layer, and $e_{c,n,i}^{(1)}$ indicates the influence value of the 1st-order feature $i$ in the node $n$ for the class $c$. Analogously, the influence of a 2nd-order feature is defined as follows:
        \begin{align}
            \begin{split}
                \textbf{a}_{n,i,j}^{(2)} &= \left( x_{n,i}^{(0)} \textbf{w}_{i}^{(1)} \odot x_{n,j}^{(0)} \textbf{w}_{j}^{(1)} \right) \textbf{W}^{(2)} \\
                &= \left( \textbf{a}_{n,i}^{(1)} \odot \textbf{a}_{n,j}^{(1)} \right) \textbf{W}^{(2)},
            \end{split}\\
            e_{c,n,i,j}^{(2)} &= \textbf{a}_{n,i,j}^{(2)} \textbf{w}_{u:2 \cdot u, c}^{(\text{out})},
        \end{align}
        where $\textbf{a}_{n,i,j}^{(2)}$ is the representation vector of the 2nd-order feature $i \times j$, and $e_{c,n,i,j}$ is the influence value of the 2nd-order feature $i \times j$ for the class $c$. Therefore, we can generalize the above equations as follows:
        \begin{align}
            \textbf{a}_{n,i,...,k}^{(l)} &= \left( \textbf{a}_{n,i,...,j}^{(l-1)} \odot \textbf{a}_{n,k}^{(1)} \right) \textbf{W}^{(l)}, \\
            e_{c,n,i,...,k}^{(l)} &= \textbf{a}_{n,i,...,k}^{(l)} \textbf{w}_{l \cdot u:(l+1) \cdot u, c}^{(\text{out})},
        \end{align}
        where $e_{c,n,i,...,k}^{(l)}$ indicates the influence value of a $l^{\text{th}}$-order feature.
	
    \subsection{Joint Learning with GNNs}

        Our EFI-GNN can be jointly trained with another GNN. In this section, we introduce a joint learning approach that combines EFI-GNN and an existing GNN. Each layer of EFI-GNN explicitly learns different-order feature interactions, while each layer of an existing GNN implicitly learns higher-order feature interactions - with upper layers capturing more complex patterns than lower layers. Since both EFI-GNN and GNN capture different patterns at each layer, utilizing these rich feature interactions can improve predictive performances. The final representation matrices of both EFI-GNN and GNN are defined as follows:
        \begin{gather}
            \textbf{X}_{\text{efi}}^{(\text{out})} = \left[ \textbf{X}_{\text{efi}}^{(1)} \parallel \textbf{X}_{\text{efi}}^{(2)} \parallel ... \parallel \textbf{X}_{\text{efi}}^{(L_{\text{efi}})} \right], \\
            \textbf{X}_{\text{gnn}}^{(\text{out})} = \left[ \textbf{X}_{\text{gnn}}^{(1)} \parallel \textbf{X}_{\text{gnn}}^{(2)} \parallel ... \parallel \textbf{X}_{\text{gnn}}^{(L_{\text{gnn}})} \right],
        \end{gather}
        where $\textbf{X}_{\text{efi}}^{(\text{out})} \in \mathbb{R}^{N \times K \cdot L_{\text{efi}}}$ and $\textbf{X}_{\text{gnn}}^{(\text{out})} \in \mathbb{R}^{N \times K \cdot L_{\text{gnn}}}$ are the final representation vectors of EFI-GNN and GNN, respectively, and $L_{\text{efi}}$ and $L_{\text{gnn}}$ are the numbers of layers of EFI-GNN and GNN, respectively. The final output layer of the joint-learning method is defined as follows:
        \begin{equation}
            \hat{\textbf{Y}} = \left[ \textbf{X}_{\text{efi}}^{(\text{out})} \parallel \textbf{X}_{\text{gnn}}^{(\text{out})} \right] \textbf{W}^{(\text{out})},
        \end{equation}
        where $\textbf{W}^{(\text{out})} \in \mathbb{R}^{K \cdot (L_{\text{efi}} + L_{\text{gnn}}) \times O}$ is the trainable output matrix.
        We use the cross-entropy as the loss function defined as:
        \begin{equation}
            \mathcal{L} = -\sum_{i \in \mathcal{V}} \textbf{y}_i log \hat{\textbf{y}}_i,
        \end{equation}
        where $\mathcal{V} = \left\{v_1, v_2, ..., v_N\right\}$ is the set of nodes.

    \section{Experiments}

    \begin{figure*}
    \centering
    \includegraphics[width=0.9\linewidth]{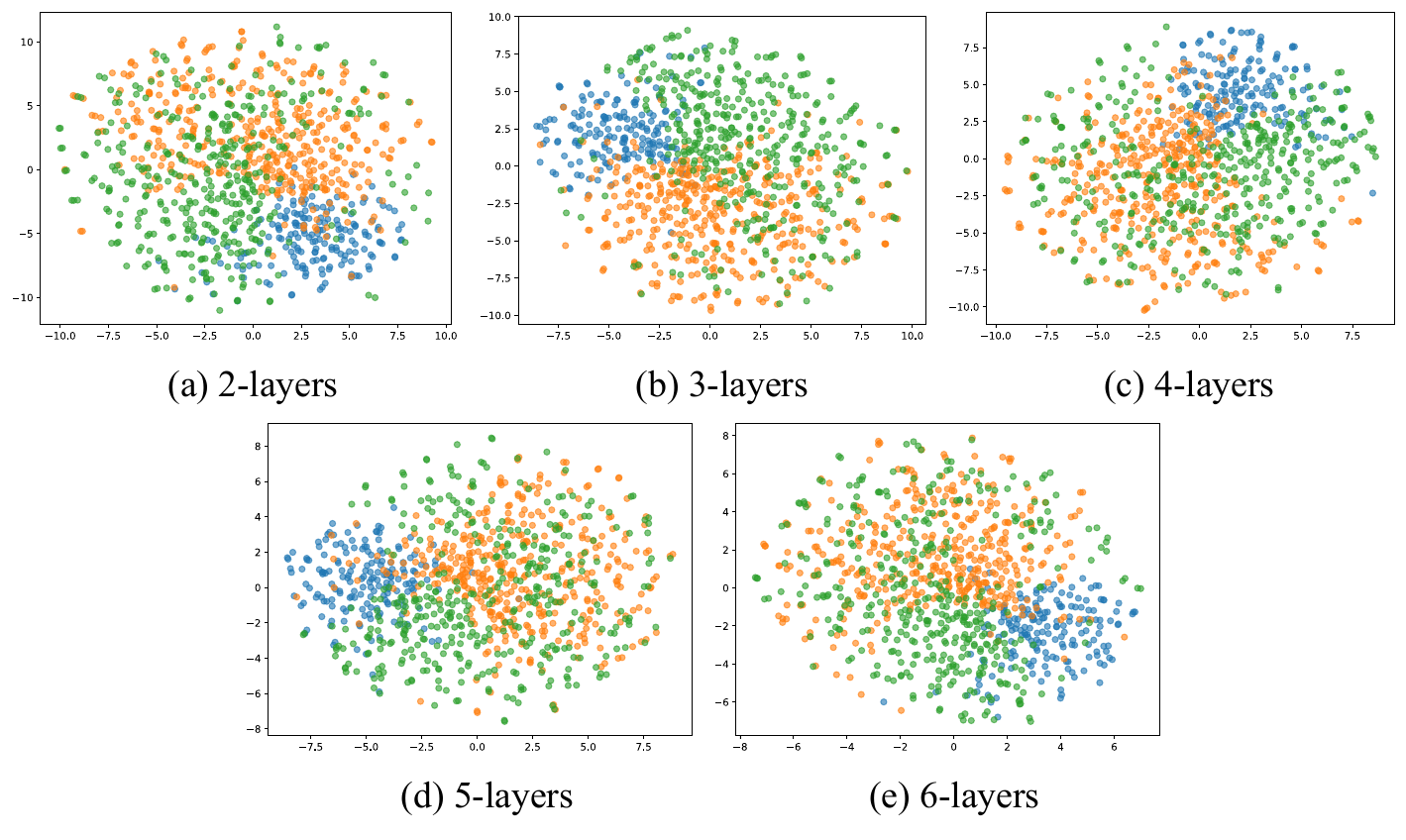}
    \caption{tSNE results of EFI-GNNs' final representations on the PubMed.}
    \label{fig:tsne}
\end{figure*}

    \begin{figure}
    \centering
    \includegraphics[width=\linewidth]{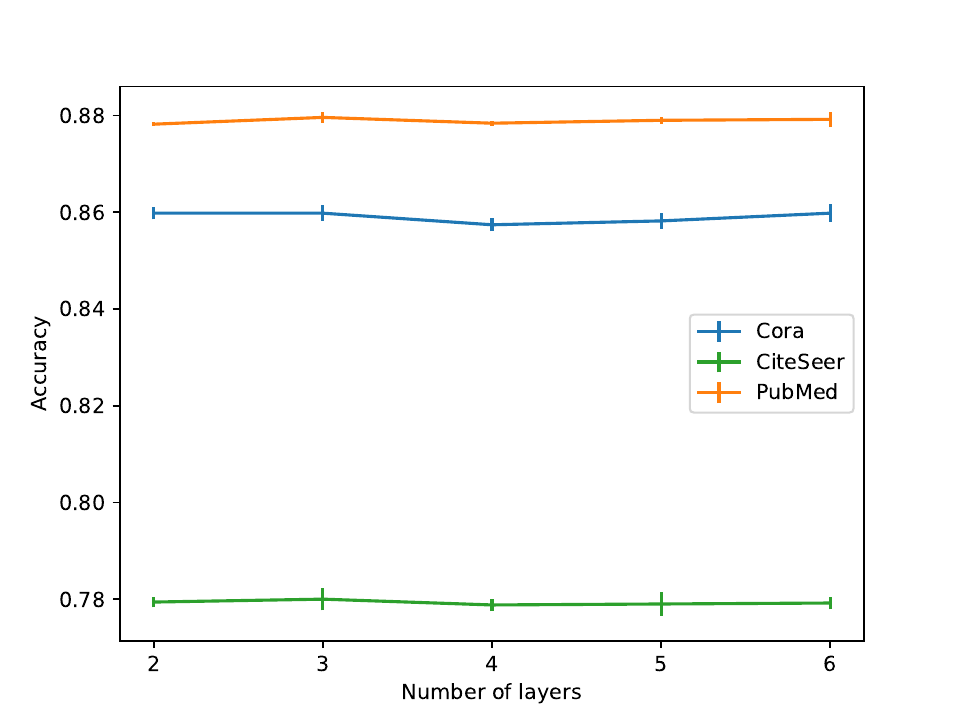}
    \caption{Ablation study on the number of layers for the Cora, CiteSeer, and PubMed.}
    \label{fig:ablation}
\end{figure}

    \subsection{Datasets}

        We compare the performance of the proposed EFI-GNN and the joint-learning methods against other leading GNNs using three citation network datasets: Cora, CiteSeer, and PubMed, \textcolor{\highlighting}{as well as the Amazon Computers \cite{shchur2018pitfalls} and Actor \cite{pei2020geom} datasets. In the Amazon, nodes indicate products in the Computers section and edges represent that two products are frequently bought together. In the Actor, each node indicates an actor, and an edge denotes the co-occurrence of two actors on the same Wikipedia page.} In addition, we conduct experiments with two open graph benchmark (OGB) datasets to explore whether combining multiple feature interactions effectively improves predictive performance. The statistics of the experimental datasets are outlined in Table~\ref{tbl:dataset_statistics}. The objectives of all datasets are to predict the classes of nodes. The obgn-mag is a heterogeneous network. Thus, we convert it to a homogeneous network by focusing solely on the (paper, cites, paper) relationships. For the Cora, CiteSeer, \textcolor{\highlighting}{Amazon, and Actor}, nodes' features are zero/one encoded vectors representing absent/present words. For the PubMed, nodes' features represent TF/IDF score vectors of the words. For the obgn-arxiv and ogbn-mag, nodes' features represent the average of word2vec vectors of the present words. All experimental datasets are publicly available. For the Cora, CiteSeer, and PubMed, we split the datasets into training/validation/test sets with the same scenario as \cite{chen2018fastgcn}. \textcolor{\highlighting}{For the Amazon and Actor, we randomly split the datasets in training/validation/test sets with 60\%/20\%/20\% ratio.} For the obgn-arxiv and ogbn-mag, we split the datasets into training/validation/test sets with the same scenario as the public OGB test.

    \subsection{Experimental Setup}
    
        We implemented the proposed EFI-GNN and joint-learning methods using PyTorch \cite{paszke2019pytorch} and PyTorch Geometric \cite{matthias2019fast}. We trained the EFI-GNN, GCN, and GCNII models on an NVIDIA GTX 1080ti with 12GB VRAM, while the remaining models, namely GAT, GATv2, and ASDGN, were trained on an NVIDIA A100 with 80GB VRAM. Training the GAT, GATv2, and ASDGN models on the GTX 1080ti for the ogbn-mag dataset is infeasible due to their high memory usage. Specifically, our PyTorch Geometric implementations of these models require 30GB, 42GB, and 18GB of VRAM, respectively.

        For all experiments, L2 penalty and dropout are applied \cite{srivastava2014dropout}. For the PubMed, ogbn-arxiv, and ogbn-mag, batch normalization \cite{ioffe2015batch} is used. To expedite the approximation speed of the models on PubMed, the residual connection \cite{he2016deep} is employed. We optimize hyperparameters, including learning rate and number of units for the Cora, CiteSeer, PubMed, \textcolor{\highlighting}{Amazon, and Actor,} using grid search. For a fair comparison, we use the same model structure and hyperparameters for ogbn-arxiv and ogbn-mag. All experimental models were trained 200 epochs for Cora, CiteSeer, and PubMed, while they were trained 1000 epochs for the ogbn-arxiv and ogbn-mag. Cross entropy is used as the loss function, and trainable parameters are optimized with Adam optimizer. Table~\ref{tbl:hparam1} shows the hyperparameter settings for experimental datasets.

    \subsection{Performance Comparison}
    
        We compare EFI-GNN with GCN \cite{kipf2017semi}, FastGCN \cite{chen2018fastgcn}, GraphSage \cite{hamilton2017inductive}, ASGCN \cite{huang2018adaptive}, GCNII \cite{chen2020simple}, GAT \cite{velivckovic2018graph}, GATv2 \cite{brody2022attentive}, and ASDGN \cite{gravina2023anti} using the Cora, CiteSeer, PubMed, Amazon, and Actor. Table~\ref{tbl:cite_acc} shows the test set accuracies for experimental models. In the \textit{"reported"} row of Table~\ref{tbl:cite_acc}, we have borrowed the reported performances of GCN, FastGCN, GraphSage, ASGCN, and GCNII from their original papers. In addition, we conducted our own experiments with GCN, GAT, GCNII, GATv2, and ASDGN. We measured the accuracies of the experimental models with 10 random seeds and reported the average accuracies and standard deviations. Our experimental models concatenate all layers to make predictions. Although our EFI-GNN is linear, it shows comparable performance to GCN. This demonstrates that EFI-GNN can effectively learn feature interactions on graphs. In addition, the predictive accuracies are consistently improved in the all joint-learning settings. This shows that the low-order feature interactions learned by EFI-GNN include useful information that cannot be captured in higher-order interactions.

        We conduct an additional experiment on two OGB datasets to examine whether combining multiple feature interactions is effective. In this experiment, GCN, GAT, GATv2, and ASDGN are used to learn implicit feature interactions, and EFI-GNN is used to learn explicit feature interactions. Table~\ref{tbl:ogb_acc} shows the test accuracies of the experimental models. GCN$^*$ and EFI-GNN$^*$ indicate GCN and EFI-GNN that make predictions using only the last layer, respectively. GCN, GAT, GATv2, and ASDGN use the concatenation of all layers to make predictions. Thus, they can learn multiple implicit higher-order feature interactions. EFI-GNN also uses the concatenation of all layers to make predictions. It can learn multiple various-order feature interactions. All joint learning methods can simultaneously learn multiple implicit higher-order and explicit various-order feature interactions. For a fair comparison, we use GCN \& GCN, which is a joint-learning method of two GCNs, and EFI-GNN \& EFI-GNN, which is the joint-learning method of two EFI-GNNs. They have the same number of trainable parameters as GCN \& EFI-GNN. GCN and EFI-GNN outperform GCN$^*$ and EFI-GNN$^*$, respectively. This shows that simultaneously learning multiple feature interactions is effective for predictive performance. Furthermore, GCN \& EFI-GNN outperforms GCN \& GCN and EFI-GNN \& EFI-GNN, and the performance of other joint learning methods also consistently improved. This demonstrates that simultaneously learning both explicit and implicit feature interactions is more effective than learning only explicit or implicit ones.

    \subsection{Ablation Study}

        Deep spectral graph neural networks may cause the over-smoothing problem. To investigate whether EFI-GNN faces the same problem, we have assessed the performance of EFI-GNN on the Cora, CiteSeer, and PubMed while varying the number of layers. Fig. \ref{fig:ablation} shows the performance changes as the number of layers changes. In the experiment, we have observed no significant changes in performance. In Fig. \ref{fig:tsne}, we have visualized the tSNE results of the EFI-GNNs’ final representations on the PubMed dataset. Fig. \ref{fig:tsne} shows that EFI-GNN does not suffer from over-smoothing problems. This stability can be caused by the jumping knowledge mechanism and residual connections of EFI-GNN. Some previous works have demonstrated that the jumping knowledge mechanism and residual connections help avoid the over-smoothing problem \cite{xu2018representation,li2019deepgcns,li2020deepergcn,chen2020simple}.

    \begin{figure}[h]

    \centering
    
    \begin{subfigure}{\linewidth}
        \centering
        \includegraphics[width=0.7\linewidth]{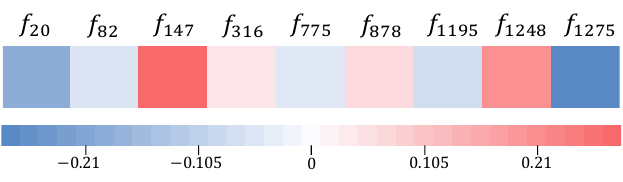}
        \caption{1st-order features}	
    \end{subfigure}
    \begin{subfigure}{\linewidth}
        \centering
        \includegraphics[width=0.7\linewidth]{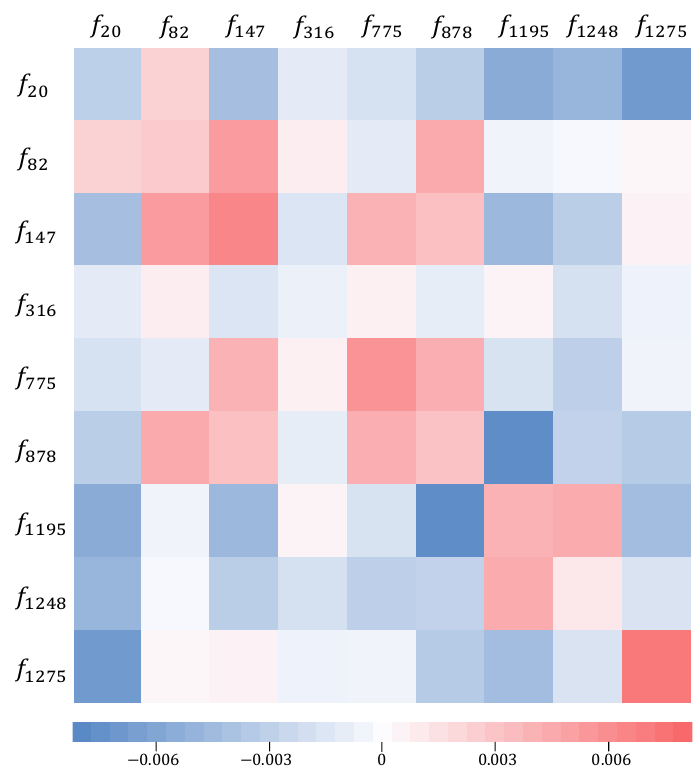}	
        \caption{2nd-order features}
    \end{subfigure}
    
    \caption{Influences of 1st- and 2nd-order features}
    
    \label{fig:feature_effect}
    
\end{figure}

    \subsection{Visualization of Feature Interactions}

        Our EFI-GNN is intrinsically interpretable, allowing us to discern the influences of features from a trained model. Fig.~\ref{fig:feature_effect} illustrates the influences of both 1st- and 2nd-order features on node $1$ with respect to class $1$ in the Cora dataset. Since the node has nine active features, we visualized the influences of these particular features. In the figure, red cells represent features with positive influences, while blue cells represent features with negative influences. Positive features increase the probability of the node label being $1$, whereas negative features decrease the probability.

        \begin{figure}[t!]

    \centering
    
    \begin{subfigure}{0.5\linewidth}
        \centering
        \includegraphics[width=\linewidth]{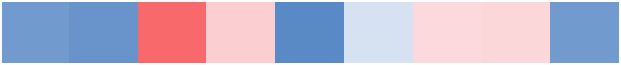}
        \caption{Units 32}
    \end{subfigure}\\
    
    \begin{subfigure}{0.5\linewidth}
        \centering
        \includegraphics[width=\linewidth]{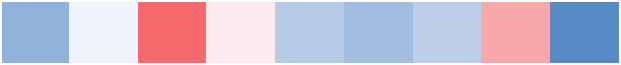}
        \caption{Units 64}
    \end{subfigure}\\
    
    \begin{subfigure}{0.5\linewidth}
        \centering
        \includegraphics[width=\linewidth]{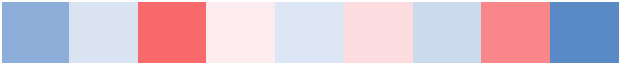}
        \caption{Units 128}
    \end{subfigure}\\
    
    \begin{subfigure}{0.5\linewidth}
        \centering
        \includegraphics[width=\linewidth]{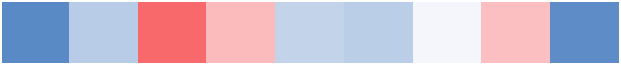}
        \caption{Units 256}
    \end{subfigure}\\
    
    \begin{subfigure}{0.5\linewidth}
        \centering
        \includegraphics[width=\linewidth]{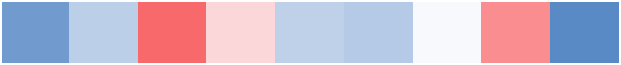}
        \caption{Units 512}
    \end{subfigure}
    
    \caption{first-order feature interactions with the different number of units}
    
    \label{fig:units}
    
\end{figure}

        \begin{figure}[t!]

\centering

    \begin{subfigure}{0.5\linewidth}
        \centering
        \includegraphics[width=\linewidth]{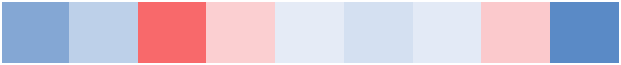}
        \caption{Layers 1}
    \end{subfigure}\\
    
    \begin{subfigure}{0.5\linewidth}
        \centering
        \includegraphics[width=\linewidth]{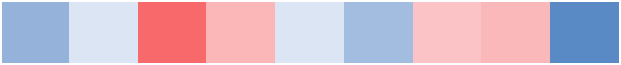}
        \caption{Layers 2}
    \end{subfigure}\\
    
    \begin{subfigure}{0.5\linewidth}
        \centering
        \includegraphics[width=\linewidth]{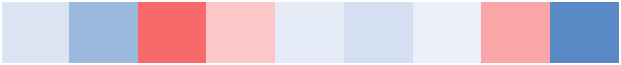}
        \caption{Layers 3}
    \end{subfigure}\\
    
    \begin{subfigure}{0.5\linewidth}
        \centering
        \includegraphics[width=\linewidth]{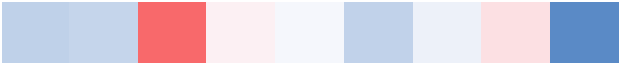}
        \caption{Layers 4}
    \end{subfigure}\\
    
    \begin{subfigure}{0.5\linewidth}
        \centering
        \includegraphics[width=\linewidth]{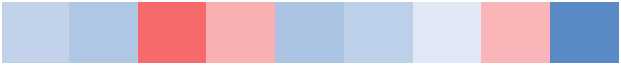}
        \caption{Layer 5}
    \end{subfigure}
    
    \caption{first-order feature interactions with the different number of layers}
    
    \label{fig:layers}
    
\end{figure}

        \begin{figure}[t!]

    \centering
    
    \begin{subfigure}{0.5\linewidth}
        \centering
        \includegraphics[width=\linewidth]{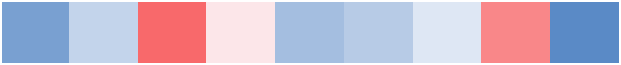}
        \caption{Seed 1}
    \end{subfigure}\\
    
    \begin{subfigure}{0.5\linewidth}
        \centering
        \includegraphics[width=\linewidth]{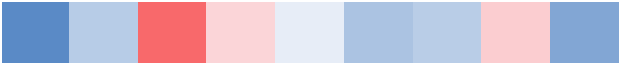}
        \caption{Seed 2}
    \end{subfigure}\\
    
    \begin{subfigure}{0.5\linewidth}
        \centering
        \includegraphics[width=\linewidth]{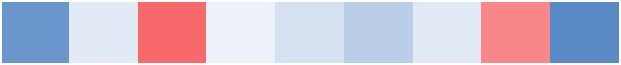}
        \caption{Seed 3}
    \end{subfigure}\\
    
    \begin{subfigure}{0.5\linewidth}
        \centering
        \includegraphics[width=\linewidth]{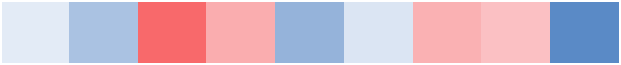}
        \caption{Seed 4}
    \end{subfigure}\\
    
    \begin{subfigure}{0.5\linewidth}
        \centering
        \includegraphics[width=\linewidth]{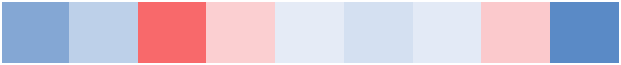}
        \caption{Seed 5}
    \end{subfigure}
    
    \caption{first-order feature interactions with the different random seeds}
    
    \label{fig:seeds}
    
\end{figure}
        \begin{figure}[t!]

    \centering
    
    \begin{subfigure}{0.4\linewidth}
        \centering
        \includegraphics[width=\linewidth]{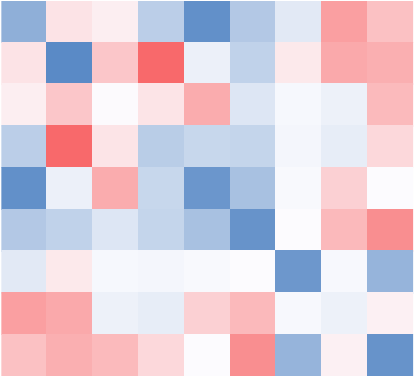}
        \caption{Units 32}
    \end{subfigure}
    \begin{subfigure}{0.4\linewidth}
        \centering
        \includegraphics[width=\linewidth]{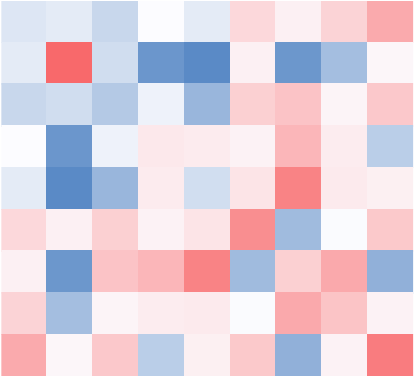}
        \caption{Units 64}
    \end{subfigure}
    \begin{subfigure}{0.4\linewidth}
        \centering
        \includegraphics[width=\linewidth]{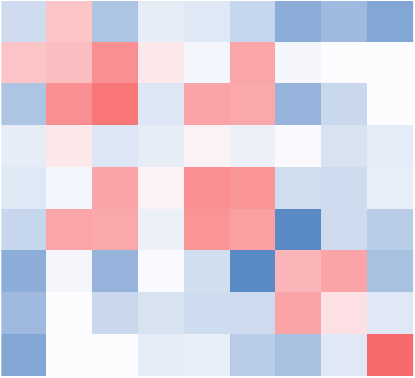}
        \caption{Units 128}
    \end{subfigure}
    \begin{subfigure}{0.4\linewidth}
        \centering
        \includegraphics[width=\linewidth]{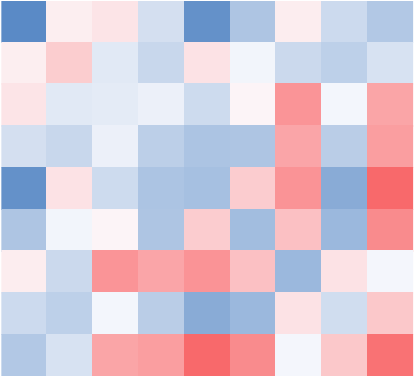}
        \caption{Units 256}
    \end{subfigure}
    \begin{subfigure}{0.4\linewidth}
        \centering
        \includegraphics[width=\linewidth]{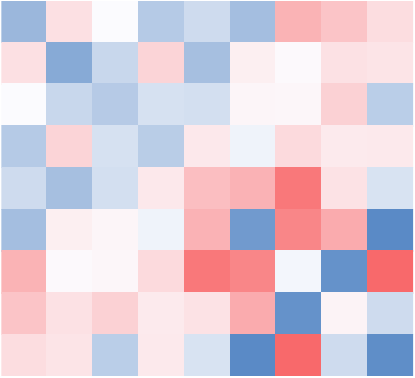}
        \caption{Units 512}
    \end{subfigure}
    
    \caption{second-order feature interactions with the different number of units}
    
    \label{fig:units2}
    
\end{figure}

		We demonstrated the interpretability of our EFI-GNN by visualizing its feature interactions. In this section, we additionally provide visualizations of 1st- and 2nd-order feature interactions across various hyper-parameter settings. Fig. \ref{fig:units}, Fig. \ref{fig:layers}, and Fig. \ref{fig:seeds} show the 1st-order feature interactions for different numbers of units, layers, and random seeds, respectively. As shown in the figures, the 1st-order interactions have a consistent trend across different hyper-parameter settings. It indicates that EFI-GNN pays attention to the same features and shows its reliability. On the other hand,  Fig. \ref{fig:units2} shows the 2nd-order feature interactions across different numbers of units. Unlike the consistent trends observed in the 1st-order interactions, it is hard to find any patterns in the 2nd-order interactions. We conjecture that the node aggregation in EFI-GNN layer causes this problem. This aggregation may make it more challenging to capture stable patterns of feature interactions. Our future work is to find out why this phenomenon is caused.

    \section{Conclusion}

    In this paper, we proposed EFI-GNN, a novel GNN architecture. EFI-GNN is a linear and interpretable model that can explicitly learn arbitrary-order feature interactions on graphs. EFI-GNN showed comparable performance to GCN, and combining EFI-GNN with another GNN further enhanced predictive performance. In addition, we visualized the influences of the 1st- and 2nd-order features through heatmaps. We believe that EFI-GNN is a pioneering work for learning feature interactions on graph-structured data.

    The major limitations of EFI-GNN are as follows: 1) Interpreting feature interactions more than 2nd-order is challenging due to neighbor aggregation. 2) EFI-GNN can only deal with homogeneous graphs. 3) Analysis of EFI-GNN behavior on homophilic and heterophilic graphs should be conducted. Based on these limitations, our avenues for future works include: 1) extending EFI-GNN to other domains where feature interactions play an important role, such as recommender systems; 2) incorporating sophisticated feature interaction learning techniques into EFI-GNN; 3) deploying EFI-GNN in high-stakes domains that necessitate interpretable models; and 4) applying EFI-GNN to other types of graphs such as heterogeneous or dynamic graphs.

    \nocite{*}
    \bibliographystyle{plain}
    \bibliography{references.bib}

\begin{thebibliography}{10}

\bibitem{berg2017graph}
Rianne van~den Berg, Thomas~N Kipf, and Max Welling.
\newblock {Graph Convolutional Matrix Completion}.
\newblock {\em arXiv preprint arXiv:1706.02263}, 2017.

\bibitem{brody2022attentive}
Shaked Brody, Uri Alon, and Eran Yahav.
\newblock {How Attentive are Graph Attention Networks?}
\newblock In {\em International Conference on Learning Representations (ICLR)},
  2022.

\bibitem{chen2018fastgcn}
Jie Chen, Tengfei Ma, and Cao Xiao.
\newblock {FastGCN: Fast Learning with Graph Convolutional Networks via
  Importance Sampling}.
\newblock In {\em International Conference on Learning Representations (ICLR)},
  2018.

\bibitem{chen2020simple}
Ming Chen, Zhewei Wei, Zengfeng Huang, Bolin Ding, and Yaliang Li.
\newblock {Simple and Deep Graph Convolutional Networks}.
\newblock In {\em International Conference on Machine Learning}, 2020.

\bibitem{chen2023learnable}
Zhaoliang Chen, Lele Fu, Jie Yao, Wenzhong Guo, Claudia Plant, and Shiping
  Wang.
\newblock {Learnable Graph Convolutional Network and Feature Fusion for
  Multi-View Learning}.
\newblock {\em Information Fusion}, 95:109--119, 2023.

\bibitem{cheng2016wide}
Heng-Tze Cheng, Levent Koc, Jeremiah Harmsen, Tal Shaked, Tushar Chandra,
  Hrishi Aradhye, Glen Anderson, Greg Corrado, Wei Chai, Mustafa Ispir, Rohan
  Anil, Zakaria Haque, Lichan Hong, Vihan Jain, Xiaobing Liu, and Hemal Shah.
\newblock {Wide \& Deep Learning for Recommender Systems}.
\newblock In {\em Proceedings of the 1st Workshop on Deep Learning for
  Recommender Systems}, 2016.

\bibitem{cho2014learning}
Kyunghyun Cho, Bart Van~Merri{\"e}nboer, Caglar Gulcehre, Dzmitry Bahdanau,
  Fethi Bougares, Holger Schwenk, and Yoshua Bengio.
\newblock {Learning Phrase Representations using RNN Encoder-Decoder for
  Statistical Machine Translation}.
\newblock In {\em Proceedings of the 2014 Conference on Empirical Methods in
  Natural Language Processing (EMNLP)}, 2014.

\bibitem{ding2019feature}
Kaize Ding, Yichuan Li, Jundong Li, Chenghao Liu, and Huan Liu.
\newblock {Feature Interaction-aware Graph Neural Networks}.
\newblock {\em arXiv preprint arXiv:1908.07110}, 2019.

\bibitem{fan2019graph}
Wenqi Fan, Yao Ma, Qing Li, Yuan He, Eric Zhao, Jiliang Tang, and Dawei Yin.
\newblock {Graph Neural Networks for Social Recommendation}.
\newblock In {\em The World Wide Web Conference}, 2019.

\bibitem{matthias2019fast}
Matthias Fey and Jan~Eric Lenssen.
\newblock {Fast Graph Representation Learning with PyTorch Geometric}.
\newblock In {\em ICLR 2019 Workshop, Representation Learning on Graphs and
  Manifolds}, 2019.

\bibitem{gravina2023anti}
Alessio Gravina, Davide Bacciu, and Claudio Gallicchio.
\newblock {Anti-Symmetric DGN: A Stable Architecture for Deep Graph Networks}.
\newblock In {\em International Conference on Learning Representations (ICLR)},
  2023.

\bibitem{guo2017deepfm}
Huifeng Guo, Ruiming Tang, Yunming Ye, Zhenguo Li, and Xiuqiang He.
\newblock {DeepFM: A Factorization-Machine Based Neural Network for CTR
  Prediction}.
\newblock In {\em Proceedings of the 26th International Joint Conference on
  Artificial Intelligence}, 2017.

\bibitem{hamilton2017inductive}
Will Hamilton, Zhitao Ying, and Jure Leskovec.
\newblock {Inductive Representation Learning on Large Graphs}.
\newblock In {\em Advances in Neural Information Processing Systems}, 2017.

\bibitem{he2016deep}
Kaiming He, Xiangyu Zhang, Shaoqing Ren, and Jian Sun.
\newblock {Deep Residual Learning for Image Recognition}.
\newblock In {\em Proceedings of the IEEE Conference on Computer Vision and
  Pattern Recognition (CVPR)}, 2016.

\bibitem{huang2018adaptive}
Wenbing Huang, Tong Zhang, Yu~Rong, and Junzhou Huang.
\newblock {Adaptive Sampling Towards Fast Graph Representation Learning}.
\newblock In {\em Advances in Neural Information Processing Systems}, 2018.

\bibitem{ioffe2015batch}
Sergey Ioffe and Christian Szegedy.
\newblock {Batch Normalization: Accelerating Deep Network Training by Reducing
  Internal Covariate Shift}.
\newblock In {\em Proceedings of the 32nd International Conference on Machine
  Learning}, 2015.

\bibitem{kim2020combining}
Minkyu Kim, Suan Lee, and Jinho Kim.
\newblock {Combining Multiple Implicit-Explicit Interactions for Regression
  Analysis}.
\newblock In {\em 2020 IEEE International Conference on Big Data (Big Data)},
  2020.

\bibitem{kipf2017semi}
Thomas~N. Kipf and Max Welling.
\newblock {Semi-Supervised Classification with Graph Convolutional Networks}.
\newblock In {\em International Conference on Learning Representations (ICLR)},
  2017.

\bibitem{li2019deepgcns}
Guohao Li, Matthias Muller, Ali Thabet, and Bernard Ghanem.
\newblock {DeepGCNs: Can GCNs Go As Deep As CNNs?}
\newblock In {\em Proceedings of the IEEE/CVF International Conference on
  Computer Vision (ICCV)}, 2019.

\bibitem{li2020deepergcn}
Guohao Li, Chenxin Xiong, Ali Thabet, and Bernard Ghanem.
\newblock {DeeperGCN: All You Need to Train Deeper GCNs}.
\newblock {\em arXiv preprint arXiv:2006.07739}, 2020.

\bibitem{li2016gated}
Yujia Li, Daniel Tarlow, Marc Brockschmidt, and Richard Zemel.
\newblock {Gated Graph Sequence Neural Networks}.
\newblock In {\em International Conference on Learning Representations (ICLR)},
  2016.

\bibitem{li2019fi}
Zekun Li, Zeyu Cui, Shu Wu, Xiaoyu Zhang, and Liang Wang.
\newblock {Fi-GNN: Modeling Feature Interactions via Graph Neural Networks for
  CTR Prediction}.
\newblock In {\em Proceedings of the 28th ACM International Conference on
  Information and Knowledge Management}, 2019.

\bibitem{li2021graphfm}
Zekun Li, Shu Wu, Zeyu Cui, and Xiaoyu Zhang.
\newblock {GraphFM: Graph Factorization Machines for Feature Interaction
  Modeling}.
\newblock {\em arXiv preprint arXiv:2105.11866}, 2021.

\bibitem{lian2018xdeepfm}
Jianxun Lian, Xiaohuan Zhou, Fuzheng Zhang, Zhongxia Chen, Xing Xie, and
  Guangzhong Sun.
\newblock {xDeepFM: Combining Explicit and Implicit Feature Interactions for
  Recommender Systems}.
\newblock In {\em Proceedings of the 24th ACM SIGKDD International Conference
  on Knowledge Discovery \& Data Mining}, 2018.

\bibitem{liu2021gcn}
Yuchen Liu, Chuanzhen Li, Han Xiao, and Juanjuan Cai.
\newblock {GCN-Int: A Click-Through Rate Prediction Model Based on Graph
  Convolutional Network Interaction}.
\newblock {\em IEEE Access}, 9:140022--140030, 2021.

\bibitem{paszke2019pytorch}
Adam Paszke, Sam Gross, Francisco Massa, Adam Lerer, James Bradbury, Gregory
  Chanan, Trevor Killeen, Zeming Lin, Natalia Gimelshein, Luca Antiga, Alban
  Desmaison, Andreas Kopf, Edward Yang, Zachary DeVito, Martin Raison, Alykhan
  Tejani, Sasank Chilamkurthy, Benoit Steiner, Lu~Fang, Junjie Bai, and Soumith
  Chintala.
\newblock {PyTorch: An Imperative Style, High-Performance Deep Learning
  Library}.
\newblock In {\em Advances in Neural Information Processing Systems}, 2019.

\bibitem{pei2020geom}
Hongbin Pei, Bingzhe Wei, Kevin Chen-Chuan Chang, Yu~Lei, and Bo~Yang.
\newblock {Geom-GCN: Geometric Graph Convolutional Networks}.
\newblock {\em arXiv preprint arXiv:2002.05287}, 2020.

\bibitem{rendle2010factorization}
Steffen Rendle.
\newblock {Factorization Machines}.
\newblock In {\em 2010 IEEE International Conference on Data Mining}, 2010.

\bibitem{rumelhart1986learning}
David~E Rumelhart, Geoffrey~E Hinton, and Ronald~J Williams.
\newblock {Learning Representations by Back-propagating Errors}.
\newblock {\em Nature}, 323(6088):533--536, 1986.

\bibitem{scarselli2008graph}
Franco Scarselli, Marco Gori, Ah~Chung Tsoi, Markus Hagenbuchner, and Gabriele
  Monfardini.
\newblock {The Graph Neural Network Model}.
\newblock {\em IEEE Transactions on Neural Networks}, 20(1):61--80, 2009.

\bibitem{shchur2018pitfalls}
Oleksandr Shchur, Maximilian Mumme, Aleksandar Bojchevski, and Stephan
  G{\"u}nnemann.
\newblock {Pitfalls of Graph Neural Network Evaluation}.
\newblock {\em arXiv preprint arXiv:1811.05868}, 2018.

\bibitem{srivastava2014dropout}
Nitish Srivastava, Geoffrey Hinton, Alex Krizhevsky, Ilya Sutskever, and Ruslan
  Salakhutdinov.
\newblock {Dropout: A Simple Way to Prevent Neural Networks from Overfitting}.
\newblock {\em Journal of Machine Learning Research}, 15(56):1929--1958, 2014.

\bibitem{tsang2020does}
Michael Tsang, Sirisha Rambhatla, and Yan Liu.
\newblock {How does This Interaction Affect Me? Interpretable Attribution for
  Feature Interactions}.
\newblock In {\em Advances in Neural Information Processing Systems}, 2020.

\bibitem{velivckovic2018graph}
Petar Veli{\v{c}}kovi{\'c}, Guillem Cucurull, Arantxa Casanova, Adriana Romero,
  Pietro Lio, and Yoshua Bengio.
\newblock {Graph Attention Networks}.
\newblock 2018.

\bibitem{wang2017deep}
Ruoxi Wang, Bin Fu, Gang Fu, and Mingliang Wang.
\newblock {Deep \& Cross Network for Ad Click Predictions}.
\newblock In {\em Proceedings of the ADKDD'17}, 2017.

\bibitem{wu2019simplifying}
Felix Wu, Amauri Souza, Tianyi Zhang, Christopher Fifty, Tao Yu, and Kilian
  Weinberger.
\newblock {Simplifying Graph Convolutional Networks}.
\newblock In {\em Proceedings of the 36th International Conference on Machine
  Learning}, 2019.

\bibitem{wu2023interpretable}
Zhihao Wu, Xincan Lin, Zhenghong Lin, Zhaoliang Chen, Yang Bai, and Shiping
  Wang.
\newblock {Interpretable Graph Convolutional Network for Multi-View
  Semi-Supervised Learning}.
\newblock {\em IEEE Transactions on Multimedia}, 2023.

\bibitem{xu2018representation}
Keyulu Xu, Chengtao Li, Yonglong Tian, Tomohiro Sonobe, Ken-ichi Kawarabayashi,
  and Stefanie Jegelka.
\newblock {Representation Learning on Graphs with Jumping Knowledge Networks}.
\newblock In {\em Proceedings of The 35th International Conference on Machine
  Learning}, 2018.

\bibitem{yang2016revisiting}
Zhilin Yang, William Cohen, and Ruslan Salakhudinov.
\newblock {Revisiting Semi-Supervised Learning with Graph Embeddings}.
\newblock In {\em Proceedings of The 33rd International Conference on Machine
  Learning}, 2016.

\bibitem{yun2019graph}
Seongjun Yun, Minbyul Jeong, Raehyun Kim, Jaewoo Kang, and Hyunwoo~J Kim.
\newblock {Graph Transformer Networks}.
\newblock In {\em Advances in Neural Information Processing Systems}, 2019.

\bibitem{zhang2020inductive}
Muhan Zhang and Yixin Chen.
\newblock {Inductive Matrix Completion Based on Graph Neural Network}.
\newblock In {\em International Conference on Learning Representations (ICLR)},
  2020.

\bibitem{zheng2021graph}
Yongsen Zheng, Pengxu Wei, Ziliang Chen, Yang Cao, and Liang Lin.
\newblock {Graph-Convolved Factorization Machines for Personalized
  Recommendation}.
\newblock {\em IEEE Transactions on Knowledge and Data Engineering}, 2021.

\end{thebibliography}
    
    \newpage
    \begin{IEEEbiography}[{\includegraphics[width=1in,height=1.25in,clip,keepaspectratio]{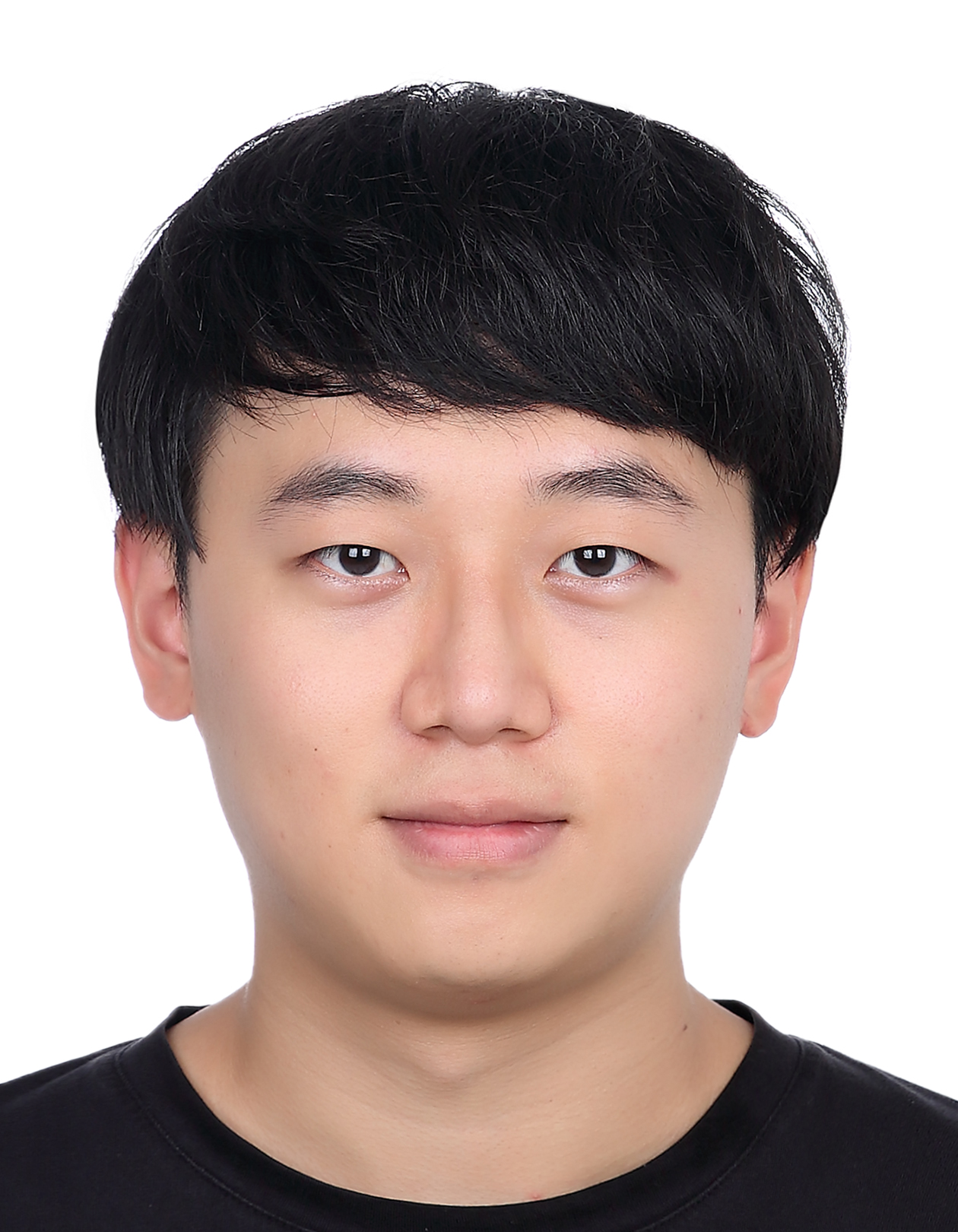}}]
    {Minkyu Kim} received the B.S. degree in computer science and the M.S. degree in big data medical convergence from Kangwon National University, Chuncheon, Republic of Korea, in 2020 and 2022, respectively.
    
    From 2022 to present, he is a Researcher at Ziovision Inc., Chuncheon, Republic of Korea.
    His research interest includes clinical decision support systems, recommender systems, graph neural networks, and explainable/interpretable AI.
\end{IEEEbiography}

\begin{IEEEbiography}[{\includegraphics[width=1in,height=1.25in,clip,keepaspectratio]{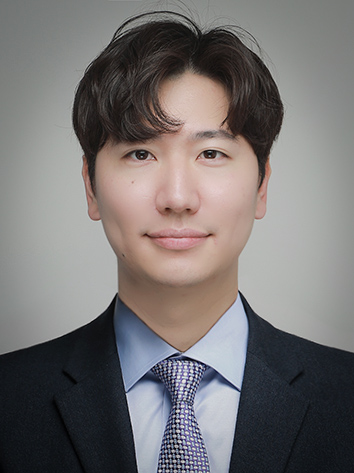}}]
    {Hyun-Soo Choi} received the B.S. degree in computer and communication engineering for the first major 
    and in brain and cognitive science for the second major from Korea University, in 2013. In 2020, 
    he also received the integrated M.S./Ph.D. degree in electrical and computer engineering from 
    Seoul National University, South Korea. From 2020 to 2021, he was a senior researcher in Vision 
    AI Labs of SK Telecom. From 2021 to 2021, he was in Kangwon National University as an assistant professor, South Korea. Since Oct 2021, he has joined 
    Ziovision as chief technical officer. Since 2023, he is working in Seoul National University of Science and Technology as an assistant professor.
\end{IEEEbiography}

\begin{IEEEbiography}[{\includegraphics[width=1in,height=1.25in,clip,keepaspectratio]{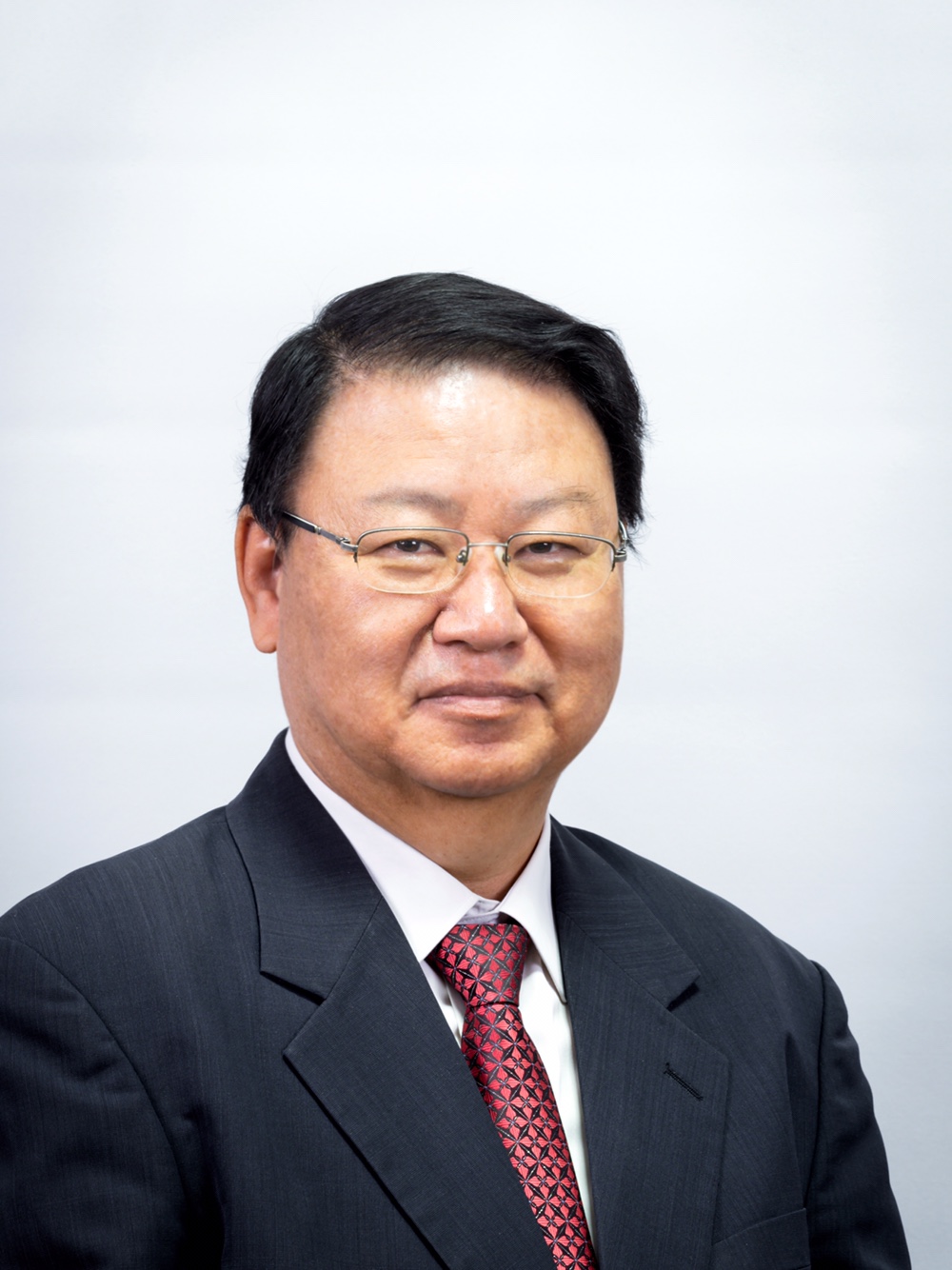}}]
    {Jinho Kim} received the B.S. degree in computer engineering from Kyungpook National University, Korea, in 1982, and the M.S. and Ph.D. degrees in computer science from Korea Advanced Institute of Science and Technology, Korea, in 1985 and 1990, respectively.
    He is currently a professor in the Department of Computer Science, Kangwon National University, Chuncheon, Korea.
    
    He was a visiting scholar with the Department of Electrical Engineering and Computer Science, University of Michigan, Ann Arbor, MI, from 1995 to 1996, the College of Computing and Informatics, Drexel University, Philadelphia, PA, from 2003 to 2004, and the School of Computing, Georgia Institute of Technology, Atlanta, GA, from 2009 to 2010, respectively.
    His research interests include high performance database systems, Hadoop/MapReduce parallel and distributed processing technology, cloud computing, data warehousing and OLAP, big data analytics, graph neural networks, medical/healthcare data analytics.
\end{IEEEbiography}

    \EOD

\end{document}